\documentclass[sigconf,nonacm]{acmart}
\usepackage{amsmath}
\usepackage{graphicx}
\usepackage[compact]{titlesec}

\AtBeginDocument{%
  \providecommand\BibTeX{{%
    \normalfont B\kern-0.5em{\scshape i\kern-0.25em b}\kern-0.8em\TeX}}}

\setcopyright{acmcopyright}
\copyrightyear{2021}
\acmYear{2021}
\acmDOI{XX.XXXX/XXXXXXX.XXXXXXX}

\acmConference[SIGKDD '21]{SIGKDD '21: ACM SIGKDD Conference on Knowledge Discovery and Data Mining}{Aug 14--18, 2021}{Singapore}
\acmBooktitle{SIGKDD '21: ACM SIGKDD Conference on Knowledge Discovery and Data Mining,
  Aug 14--18, 2021, Singapore}
\acmPrice{XX.XX}
\acmISBN{978-1-4503-XXXX-X/18/06}

\begin{document}

\fancypagestyle{firstpagestyle}
{
   \fancyhf{}
   \setlength{\footskip}{7mm}
   \fancyfoot[L]{\footnotesize Workshop on Bayesian Causal Inference for Real World Interactive Systems, 27th SIGKDD Conference on Knowledge Discovery and Data Mining (KDD 2021), Singapore.}
}
\thispagestyle{firstpagestyle}

\title{Smooth Sequential Optimisation with Delayed Feedback}

\author{Srivas Chennu}
\email{srivas.chennu@apple.com}
\author{Jamie Martin}
\email{jamiemartin@apple.com}
\author{Puli Liyanagama}
\email{puli@apple.com}
\author{Phil Mohr}
\email{phil.mohr@apple.com}
\affiliation{%
  \institution{Apple}
  \country{}
}

\renewcommand{\shortauthors}{Chennu et al.}

\begin{abstract}
Stochastic delays in feedback lead to unstable sequential learning using multi-armed bandits. Recently, empirical Bayesian shrinkage has been shown to improve reward estimation in bandit learning. Here, we propose a novel adaptation to shrinkage that estimates smoothed reward estimates from windowed cumulative inputs, to deal with incomplete knowledge from delayed feedback and non-stationary rewards. Using numerical simulations, we show that this adaptation retains the benefits of shrinkage, and improves the stability of reward estimation by more than 50\%. Our proposal reduces variability in treatment allocations to the best arm by up to 3.8x, and improves statistical accuracy -- with up to 8\% improvement in true positive rates and 37\% reduction in false positive rates. Together, these advantages enable control of the trade-off between speed and stability of adaptation, and facilitate human-in-the-loop sequential optimisation.
\end{abstract}





\maketitle

\section{Introduction}
Sequential adaptive optimisation is an example of a reinforcement learning system that progressively adapts the allocation of treatments in tune with the observed responses to these treatments. This methodology is typically framed as the classical multi-armed bandit problem of reward learning, and finds valuable applications in many domains, including clinical medicine \cite{Press2009BanditResearch}, political and social sciences \cite{Imai2011EstimationCampaign, Benartzi2017ShouldNudging}. It is also used for large-scale optimisation in the technology industry \cite{Scott2010ABandit}. In this context, it has improved the efficiency of online systems by automatically selecting better performing alternatives with minimal manual intervention. However, in many real-world applications of this technology, human operators are involved in higher-order decision making about generating treatment candidates, tuning optimisation goals, and tracking adaptation \cite{Bakshy2018AE:Experimentation}. To support such \emph{human-in-the-loop} optimisation, the stability of adaptation is as important a goal as its efficiency. Smooth adaptation is naturally more interpretable, and enables human operators to conduct reliable hypothesis testing alongside the adaptation.

A particular challenge for achieving stable adaptation is delayed feedback. In many applications, stochastic delays in receiving feedback about treatments are common \cite{Garg2019StochasticFeedback, Vernade2020LinearFeedback}. These delays can arise for various reasons -- they can occur in actually administering the treatment once allocated, in measuring and communicating the response, and in the process of receiving and processing the response. Together, the additive consequence of such delays poses a challenge to the smoothness and stability of adaptation. The challenges posed by delayed feedback are further exacerbated when rewards are non-stationary, as is common in many practical scenarios. As we show, in these scenarios, maximum likelihood estimation (MLE) of rewards results in unstable learning behaviour, and makes hypothesis testing unreliable. The option of waiting until all the delayed feedback has been received is equally unattractive, as it would considerably slow down the pace of learning.

To help address this challenge, we propose computational improvements to recent applications of empirical Bayesian (eB) methods in this context \cite{Dimmery2019ShrinkageExperiments}. Compared to MLE, our improvements enable control of the trade-off between stability and speed of adaptation, supporting \emph{interpretable} sequential optimisation. In addition to minimising regret, we improve the accuracy of hypothesis testing. Our novel contribution is an algorithmic adaptation of eB shrinkage estimation to learn from delayed feedback, such that
\begin{itemize}
    \item treatment allocations adapt smoothly to reward estimates that change due to delayed feedback and non-stationarity.
    \item cumulative variance in estimation error is minimised in the long run.
    \item statistical power of sequential Bayesian testing of hypotheses is maximised.
\end{itemize}

\section{Related Work}
\subsection{Delayed Anonymous Feedback}
We assume that individual responses to treatment allocations are generated by stochastic generative processes that govern both the response itself, and the delay in its generation. This follows the \emph{stochastic delayed composite anonymous feedback} (SDCAF) model of bandit learning \cite{Garg2019StochasticFeedback}, which extends beyond previous work in this space \cite{Pike-Burke2018, Cesa-Bianchi2018NonstochasticFeedback, Vernade2020LinearFeedback}. It assumes that responses are individually delayed and revealed at random points in the future. We extend this model to the batched online learning scenario, where we demonstrate the value of smooth empirical Bayesian shrinkage estimation.

\subsection{Batched Bandit Learning}
We combine the SDCAF model above with batch-based learning \cite{Perchet2016BatchedProblems}. Batched learning extends the classic bandit problem to scenarios where a reinforcement learning agent pulls a batch of arms between consecutive learning updates. Batched learning has been well studied in the pure exploration setting, where the aim is to find the top-\emph{k} arms as quickly as possible while minimising exploration cost, both in the absence \cite{Jun2016TopPulls} and presence \cite{Grover2018BestFeedback} of delayed feedback.

Here, we focus on the online batched learning scenario, where the agent learns as it goes, both exploring and exploiting between consecutive updates. This is typical in the digital context \cite{Scott2010ABandit}, where individual responses are stored as they arrive and aggregated together anonymously to then update the agent, once per time interval $T$ between consecutive batched updates. We focus our analysis on the Thompson sampling agent \cite{Agrawal2012AnalysisProblem}, which is well-suited to this scenario because of its stochastic allocation strategy. It pulls arms by sampling from Bayesian posterior distributions representing current knowledge of arm rewards. It updates this knowledge once per batch, by combining the individual responses received in that batch.

Combining batched online learning with the SDCAF model above, we allow the time interval $T$ between consecutive learning updates to be much shorter than the maximal time horizon $\tau = UT$ allowed for the receipt of individually delayed responses, where $U$ is an integer number of consecutive updates. A natural consequence of this setup is that at every periodic update, the batch of data processed can include responses to treatment allocations ``spilling over'' from any of the $U$ batches in the past.

\subsection{Empirical Bayesian Shrinkage}
Empirical Bayesian estimation has been shown to be valuable for statistical learning at scale \cite{Efron2011Large-ScalePrediction}, and has recently been proposed for the batched, online sequential learning setting by Dimmery et al. \cite{Dimmery2019ShrinkageExperiments}. Specifically, the authors developed a novel application of James-Stein shrinkage to reduce reward estimation error in multi-armed bandits, and demonstrated its value for learning about the best \emph{set} of arms. They showed that the benefit of eB for reward estimation grows with the number of arms. However, as they state, their formulation specifically assumes no ``spillovers'' of responses, hence avoiding the problem of delayed feedback.

Here, we extend Dimmery et al.'s work and augment eB estimation to account for the consequences of allowing such spillover generated by delays in responses. We draw upon research into empirical Bayesian smoothing \cite{Shen1999, Maritz2018EmpiricalEdition}, which has been employed to handle reporting delays in disease epidemiology \cite{Meza2003EmpiricalMapping,McGough2020NowcastingTracking}, mirroring our delayed feedback context.

\subsection{Bayesian Hypothesis Testing}
Users of adaptive optimisation systems often also want to test hypotheses about statistically significant differences between arms, alongside achieving the core optimisation objective. However, the potential conflict between the primary reward optimisation goal needs to be balanced against the need for statistical power to achieve this secondary hypothesis testing goal. As a further novel contribution of this work, we address this need by combining eB optimisation with sequential hypothesis testing using Bayes factors \cite{Schonbrodt2017SequentialDifferences}. This framework combines bandit learning with Bayesian hypothesis testing to allow users to interrogate differences between arms, at any point during the sequential optimisation process.

\section{Contribution}
We develop smoothed eB models for bandit learning that account for delayed feedback and afford control over the trade-off between speed and stability of adaptation. To the best of our knowledge, the value of smoothed eB estimation has not been previously evaluated for enabling batched online learning with delayed feedback, in particular for enabling interpretability by supporting simultaneous hypothesis testing.

Here, we bring these aspects together in a novel algorithmic contribution detailed below. We focus on practical applications in the digital context, and complement existing theoretical work in this space with numerical simulations that replicate patterns of responses and delays observed in practice. We demonstrate that our contribution improves the recent proposal by Dimmery et al. \cite{Dimmery2019ShrinkageExperiments} when handling delayed feedback, both in terms of stability of adaptive optimisation \emph{and} hypothesis testing accuracy.

\subsection{Shrinkage Estimation}
Between consecutive updates $u$ of the learning agent, we randomly allocate individual treatment units $i$ to treatment arms $a_i \in {1..K}$, selecting from one of $K$ treatment arms, using Thompson sampling over reward distributions. As per the delayed feedback model above, at some future update $u' \leq u + U$, we observe the $i$th unit's response $y_{i,u'}$. Consequently, at each update $u$, $n_{k,u}$ units have been allocated to arm $k$, and $y_{i:a_i=k,u}$ responses have been observed, each of which have incurred a variable, stochastic delay.

At each update $u$, we update our posterior reward distributions used for Thompson sampling. To perform this update, we employ shrinkage, a form of regularisation with many statistical benefits \cite{Gelman2012WhyComparisons}. Specifically, we build upon the approach to shrinkage estimation proposed by Dimmery et al. \cite{Dimmery2019ShrinkageExperiments}, based on the James-Stein (JS) method \cite{Efron2011Large-ScalePrediction}. Specifically, to handle delayed feedback, we aggregate over recent treatment allocations and responses. Intuitively, our proposal makes the evolution of the posteriors more gradual over consecutive updates.

More formally, given allocations $n_{k,u}$ and responses $y_{i:a_i=k,u}$ at each batch, we calculate \emph{cumulative} means $c_{k}$ and variances $v_k$ over treatment allocations and responses over $U$ most recent batches as

\begin{equation}
    c_k = \frac{1}{\sum_{u=1}^{U}n_{k,u}}\sum_{u=1}^{U}\sum_{i:a_i=k}y_{i,u}
\end{equation}

\begin{equation}
    v_k \propto \frac{1}{\sum_{u=1}^{U}\hat{\sigma}_{k,u}^2}
\end{equation}

\noindent where $\hat{\sigma}_{k,u}^2$ are the variances of the reward estimates at batch $u$. We then adapt Dimmery et al.'s formulation of the positive part JS estimator of each arm's shrunk mean $c_k^{JS}$ and variance $V_k^{JS}$, using the cumulative reward estimates:

\begin{equation}
    c_k^{JS} = \bar{c} + (1 - \xi_k)(c_k - \bar{c})
\end{equation}

\begin{equation}
    v_k^{JS} \approx (1 - \xi_k)v_k^2 + \frac{\xi_k s^2}{K} + \frac{2\xi_k^2(c_k - \bar{c})^2}{K - 3}
\end{equation}

where

\begin{equation}
    \bar{c} = \frac{\sum_{i=1}^{K} c_k}{K}
\end{equation}

and

\begin{equation}
    s^2 = \sum_{k=1}^K(c_k - \bar{c})^2.
\end{equation} 

\noindent The amount of shrinkage applied to each arm, $\xi_k$, is calculated as:

\begin{equation}
    \xi_k = \min(v_{k}^2\frac{K - 3}{s^2}, 1)
\end{equation}

With more data and lower cumulative variance $v_k$, and/or larger differences in the cumulative means $c_k$, $\xi_k$ approaches zero and eB estimation converges towards MLE.

\subsection{Stability vs. Speed of Learning}
Further, we control the stability vs. speed of convergence of treatment allocation by combining past shrunk estimates from $U$ recent updates. We calculate \emph{smoothed} mean $C_{k}^{JS}$ and variance $V_{k}^{JS}$ over cumulative JS estimates $c_k^{JS}$ and $v_k^{JS}$ by combining recent updates:

\begin{equation}
    C_{k}^{JS} = \sum_{u=1}^{U}w_u * c_{k,u}^{JS}
\end{equation}

\begin{equation}
    V_{k}^{JS} \propto \frac{1}{\sum_{u=1}^{U}w_u * v_{k,u}^{JS}}
\end{equation}

\noindent where $w_u$ is a smoothing weight assigned to the JS estimate from update $u$. The choice of $w_u$ allows control over the trade-off between speed and stability of convergence, as it allows us to control the influence of past estimates on the future ones. We develop alternatives for choosing $w_u$ that represent specific points along that trade-off. For example, setting

\begin{equation}
    w_u = 
    \left\{\begin{array}{ll}
    1 & u = 1, \\
    0 & \textit{otherwise} \\
    \end{array}\right\}
\end{equation}

\noindent speeds up learning by only using the most recent JS estimate. However, this results in an unstable reinforcement learning loop, manifesting as oscillations in treatment allocations proportions $n_{k,u}$ from one update to the next, especially early in the learning process. This undesirable behaviour occurs because the delayed feedback model leads to spillover in the responses $y_{a_i,u}$ observed, causing alternating over- and under-estimation, even with shrinkage.

Alternatively, a \emph{uniform} smoothing approach specified by $w_u = \frac{1}{U}$ assigns equal weights to recent JS estimates, thereby gaining smooth, stable allocation behaviour by averaging over the spillover, but at the cost of potentially underestimating rewards. A \emph{discounted} smoothing approach specified by $w_u = 1 - \frac{u - 1}{U}$, linearly tapers the contribution of JS estimates based on their recency, affording a middle ground solution.

\subsubsection{Stationary vs. Non-stationary Rewards}
The benefits of smoothing can be extended by controlling $U$, the length of the smoothing window. Setting $U$ to be equal to the total number of updates grows it linearly, producing asymptotic convergence, desirable under the assumption that true rewards are stationary. When they are non-stationary, and in particular \emph{piece-wise} stationary, $U$ can be bounded to a sliding window over recent updates \cite{Garivier2008OnProblems}. This bounds the maximal delay to the length of the sliding window, converting the uniform and discounted smoothing alternatives described above to their non-stationary counterparts described by Garivier et al. \cite{Garivier2008OnProblems}.

\subsection{Sequential Bayes Factors}
Given a pair of arms, $k_1$ and $k_2$, we evaluate the relative evidence for the hypotheses $H0$: that the true rewards of the arms are statistically indistinguishable and $H1$:  that the true rewards of $k_1$ is greater than $k_2$ by at least a certain minimum detectable effect size.

At each update $u$, we calculate this evidence in the form of $\log_{10}$ sequential Bayes factors \cite{Schonbrodt2017SequentialDifferences}, which are interpreted as per \cite{Kass1995BayesFactors}. Specifically, given the smoothed JS estimates defined as above, we calculate sequential Bayes factors (sBF) as

\begin{equation}
    BF = \frac{p(\lvert C_{k_1}^{JS} - C_{k_2}^{JS} \rvert \;\vert H1)}
              {p(\lvert C_{k_1}^{JS} - C_{k_2}^{JS} \rvert \;\vert H0)}
\end{equation}

$H0$ and $H1$ can be initialised with prior expectations about the difference between the true rewards of $k_1$ and $k_2$ if available. If not, we initialise them with Gaussian noise priors \cite{Dienes2014}.

\section{Simulations}

\begin{figure*}[!htbp]
  \centering
  \includegraphics[width=\textwidth]{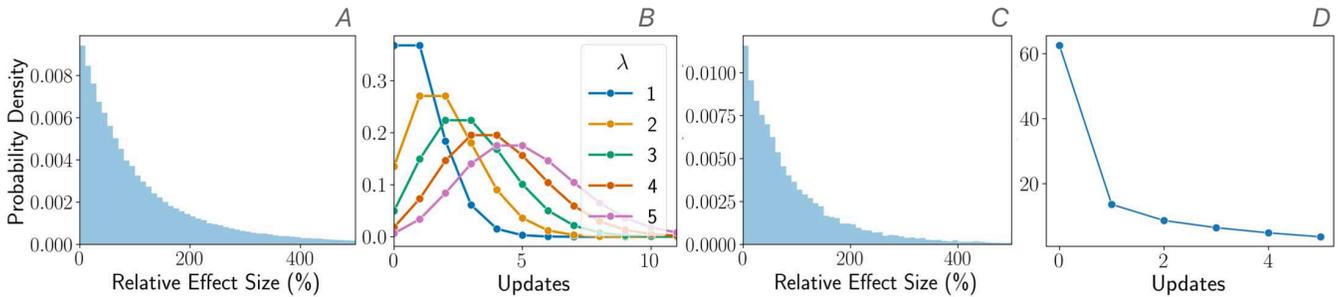}
  \caption{\textbf{Relative effect sizes and delays}. Panels A and B plot relative effect sizes and Poissonian delay distributions in synthetic simulations.  Panels C and D plot the same in realistic simulations.}
  \label{Sims}
\end{figure*}

We used numerical simulations to measure the performance of our smoothed eB alternatives to vanilla eB \cite{Dimmery2019ShrinkageExperiments} and maximum likelihood estimation (MLE). We compared performance in three simulation contexts:

\begin{itemize}
    \item with \emph{synthetic} arm rewards and response delays.
    \item under \emph{stationary} or \emph{non-stationary} reward conditions.
    \item with \emph{real} rewards and delays observed at our company.
\end{itemize}

\subsection{Synthetic Stationary Rewards}
Each simulation consisted of 500 repeated trials. Each trial consisted of a sequence of 300 learning updates of the agent. At each such update $u$, we allocated treatments to 1000 units and calculated smoothed estimates from the received responses. We assumed that true rewards were stationary, and hence, combined JS estimates from all updates between $1$--$u$. These JS estimates themselves are calculated from cumulative treatment allocations and responses between updates $1$--$u$. At the beginning of each trial, we created 15 treatment arms, with randomly initialised binary rewards $p$, unknown to the learning agent. In synthetic simulations, we sampled from a beta distribution with parameters $\alpha=3,\beta=80$. Fig. \ref{Sims}A plots the distribution of true relative effect sizes, between all simulated arm pairs and across all trials.

We also randomly initialised each arm in a trial with a probabilistic delay also unknown to the agent, modelled as Poisson distributions $P$ with $\lambda$ in $1$--$5$, illustrated in Fig. \ref{Sims}B. Given an arm $k$ with a randomly initialised delay $\lambda_k$, individual treatment responses $y_i$ generated by the arm would be stored in a buffer, and provided to the agent after a delay of $l \sim P(\lambda_k)$ updates. Hence, in a simulation run, arms would not only have unequal reward probabilities, they would also have unequal delays.

\subsection{Synthetic Non-stationary Rewards}
We simulated a simple non-stationary (piece-wise stationary) scenario where the true rewards are randomly interchanged at a single change point, by repeating the synthetic simulations above with two key differences:

\begin{itemize}
    \item The length of the smoothing window $U$ was bounded to 50 updates, implementing a sliding window \cite{Garivier2008OnProblems} much larger than the maximal response delay (Fig. \ref{Sims}B).
    \item At update 100 in each trial, unknown to the agent, the true rewards of the arms were randomly interchanged.
\end{itemize}

\subsection{Realistic Stationary Rewards}
In realistic simulations, we instead sampled from distributions of real rewards and delays in treatment responses observed at our company. As Figs. \ref{Sims}C and \ref{Sims}D show, relative effect sizes and delays observed in reality were similar to the synthetic context.

In both synthetic and realistic simulations, we set the minimum detectable effect size to be $0.01$. Across all simulation trials, approximately 55\% of true differences between rewards were above this threshold, and were used to calculate the true positive rate, with the remaining used to calculate the false positive rate.

\section{Results}
We demonstrate improved stability of our two smoothed eB alternatives -- eB with uniform smoothing (\emph{useB}) and discounted smoothing (\emph{dseB}) -- when compared to vanilla eB without smoothing (\emph{veB} \cite{Dimmery2019ShrinkageExperiments}) and standard \emph{MLE} without eB. We first report results from simulations with stationary synthetic rewards and delays, followed by non-stationary rewards. We then present results from simulations based on realistic rewards and delays observed at our company.

\subsection{Treatment Allocation}

\begin{figure*}[!htbp]
  \centering
  \includegraphics[width=\textwidth]{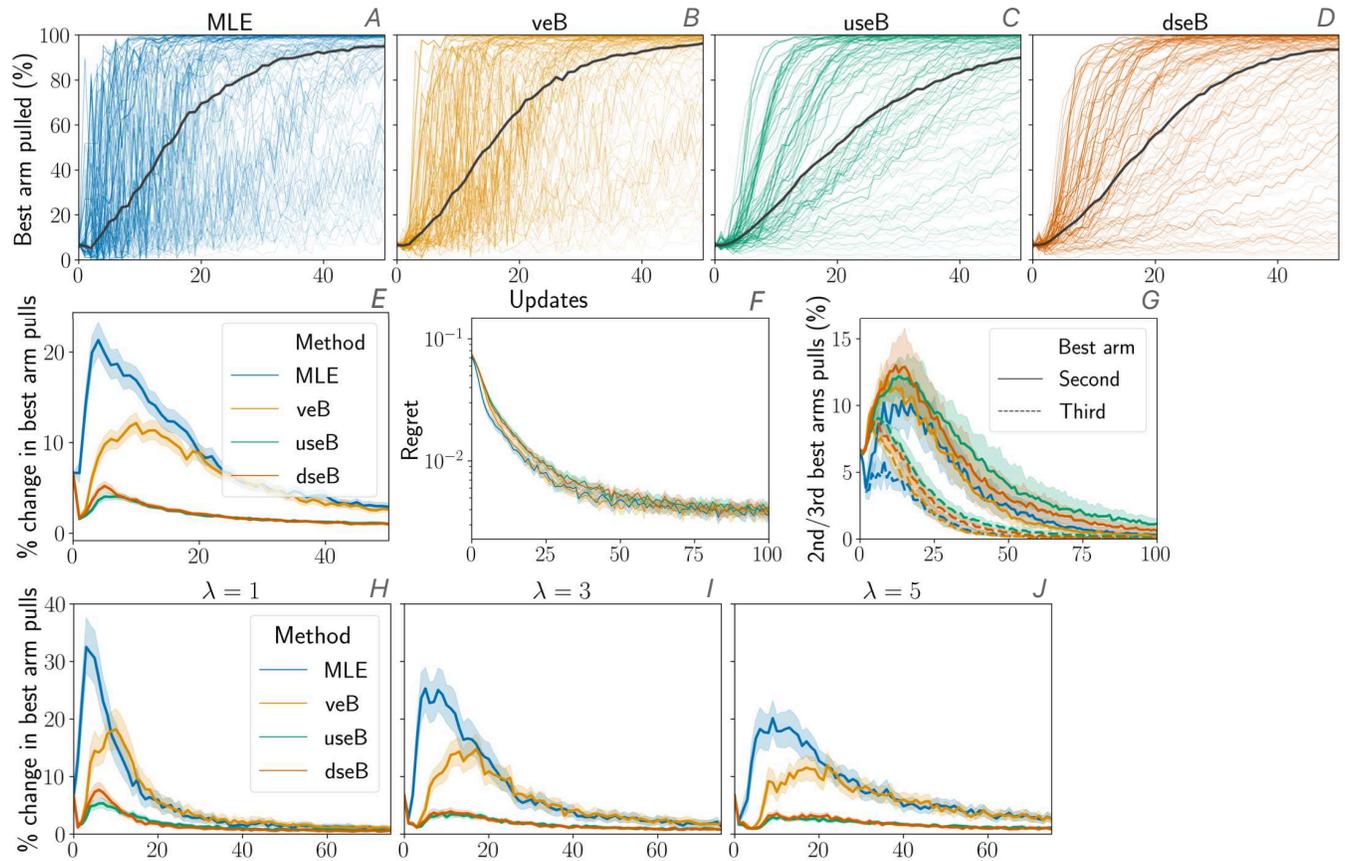}
  \caption{\textbf{Smoothing improves the stability of adaptive optimisation under delayed feedback}. In panels A-D, coloured lines represent a sample of 100 individual trials, and plot the percentage of allocations to the best arm at each update. Thickness of each line represents the effect size between the best arm and the next best one in the corresponding trial. Black line represents median allocation percentage to the best arm, over all 500 trials. Panel E plots the change in allocations to the best arm over consecutive updates, averaged over all trials, while panel F plots regret. In panel G, each line plots median percentage of pulls of 2nd and 3rd best arms. Panels H--J plot change in allocations to the best arm for different values of the $\lambda$ parameter of the Poissonian delay distribution. Shaded areas in panels E--J represent 95\% confidence intervals, computed over 1000 bootstrap iterations.}
  \label{Pulls}
\end{figure*}

Fig. \ref{Pulls} compares the adaptation of treatment allocations, identifying relevant differences between the alternatives. Allocations adapted to the best arm in each trial, as function of how much better the best arm's reward was, compared to the next best one. However, due to the impact of delayed feedback, both MLE and veB display unstable allocations to the best arm. In particular, this instability resulted in large oscillations that were especially prominent in the first 25 updates. In contrast, the smoothed variants of eB -- useB and dseB -- display a slower but stable pattern of adaptation that converges towards MLE with the reduction of shrinkage over updates. This observation is highlighted in Fig. \ref{Pulls}E, which shows the average change (across trials) in allocation to the best arm from one update to the next. useB and dseB evidenced the most stable pattern of adaptation, facilitating better interpretation of system behaviour. Though veB was better than MLE, both were unstable across consecutive updates, with a max of 3.8x and 2.6x more instability relative to the smoothed eB alternatives, respectively. Though our approach to smoothing made adaptation to the best arm more stable, it did so without a major increase in overall regret in allocations, as shown in Fig. \ref{Pulls}F.

Fig. \ref{Pulls}G brings out a further aspect of our smooth eB alternatives. They spend more time, especially in early updates, exploring the second and third best arms more often. They adapt to allocate treatments to the best performing \emph{set} of arms, rather than just the single \emph{best} arm. In other words, they naturally manage the \emph{explore-exploit} trade-off fundamental to reinforcement learning, without the need for a separate $\epsilon$ parameter. Importantly, as evident in the behaviour of the veB alternative, such exploratory behaviour is lost due to instabilities in delayed feedback. Hence, the smoothed alternatives we propose `recover' this important advantage of eB.

\subsubsection{Stability of adaptation as a function of delay}
To provide further insights into how our proposals adapt to delayed feedback, we evaluated variability in treatment allocation to the best arm as function of the stochastic delay in receiving responses. We re-ran our simulations above, but now with the response-wise delay sampled from Poisson distribution with values of $\lambda$ that were fixed at chosen values for the entire simulation run. Figs. \ref{Pulls}H-J plot the average change in allocation to the best arm from one update to the next - the same as that plotted in Fig. \ref{Pulls}E - but for specific values of $\lambda$. Reiterating the pattern in Fig. \ref{Pulls}E, MLE and veB suffered higher instability in traffic allocation to the best arm. Further, this instability became more prolonged as the amount of delay increased. In comparison, useB and dseB showed consistent stability in allocation to the best arm, even under extended feedback delays generated when $\lambda = 5$ (see Fig \ref{Pulls}J).

\subsection{Reward Estimation}

\begin{figure*}[!htbp]
  \centering
  \includegraphics[width=\textwidth]{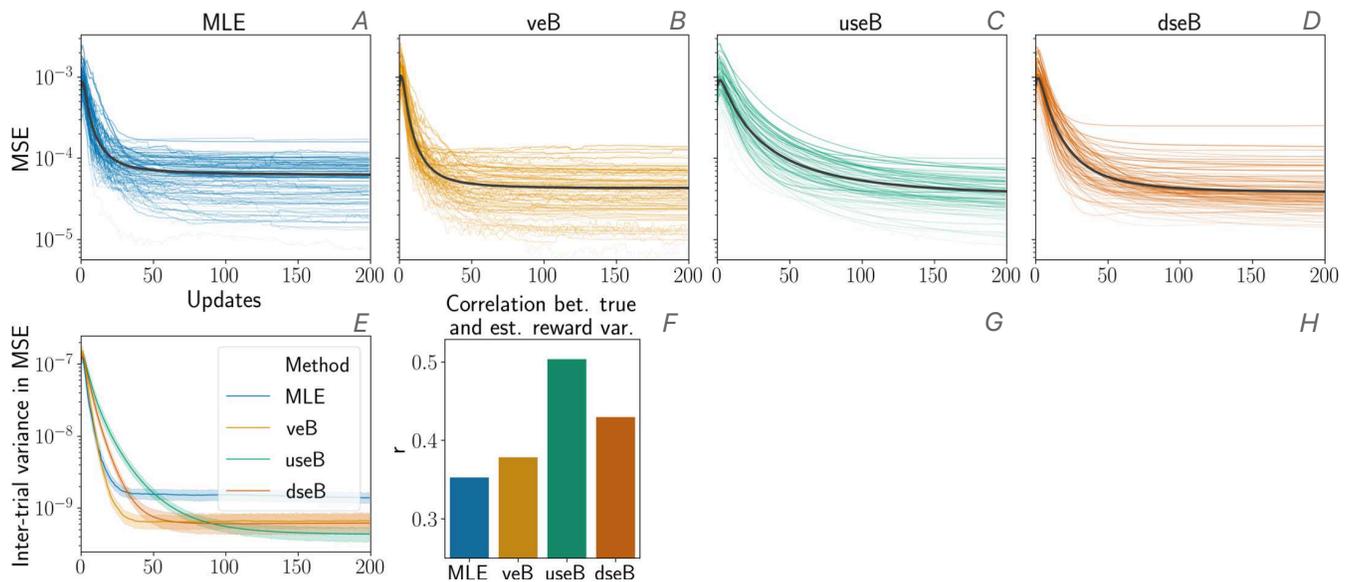}
  \caption{\textbf{Smoothing reduces variability in the estimation of delayed rewards}. In panels A-D, coloured lines represent a sample of 100 individual trials, and plot the mean squared error in reward estimation. Thickness of each line represents the variance in true rewards in the corresponding trial. Black line represents mean MSE over all 500 trials. Panel E plots the inter-trial variance in MSE. Shaded areas represent 95\% confidence intervals, computed over 1000 bootstrap iterations. Panel F plots correlation (across trials) between true and estimated rewards at the final update.}
  \label{Rewards}
\end{figure*}

Delayed feedback causes observed rewards to oscillate, even if the true rewards are stationary. The advantages of our smoothed eB approaches are underpinned by improvements to reward estimation that are robust to this noise. This improvement can be seen in Figs. \ref{Rewards}A-D. We observed lower variability in mean squared error (MSE) in reward estimation across arms with useB and dseB from one update to the next. This lower \emph{inter-update} variability in reward estimation explains the lower variance in allocations to the best arms over consecutive updates (Fig. \ref{Pulls}E).

Further, \emph{inter-trial} variance in estimation error was also reduced in the longer term, see Fig. \ref{Rewards}E. At the final update, this variance in estimation error was 52-69\% lower with eB compared to MLE, between  highlighting the general value of shrinkage. However, as can be seen in Fig. \ref{Rewards}E, variance was \emph{initially} higher with useB than MLE and veB, and less so with dseB, the middle ground option. This is to be expected, as early rewards being smoothed over are underestimates. This underestimation occurs because delays in receiving feedback mean that relatively few responses are received early on, but eventually start `piling up'. However, over subsequent updates, the value of smoothing becomes evident: the adverse effects of delays are accounted for, and inter-trial variability in reward estimation eventually stabilises at a lower level. Another insight into the behavior of our approach can be gained by measuring the across-trial \emph{correlation} between the variance in true and estimated rewards, shown in fig, \ref{Rewards}E. All eB methods, and smoothed eB methods more specifically, evinced an increased correlation. This correlation is also visually evident in the comparison of Figs. \ref{Rewards}A-D, which confirms that, with our smoothed eB alternative in particular, MSE was higher in trials with a greater proportion of truly outlying rewards. This observation mirrors a point by Dimmery et al. \cite{Dimmery2019ShrinkageExperiments}, noting that eB approaches tend to be biased against outliers.

\subsection{Hypothesis Testing}

\begin{figure*}[!htbp]
  \centering
  \includegraphics[width=\textwidth]{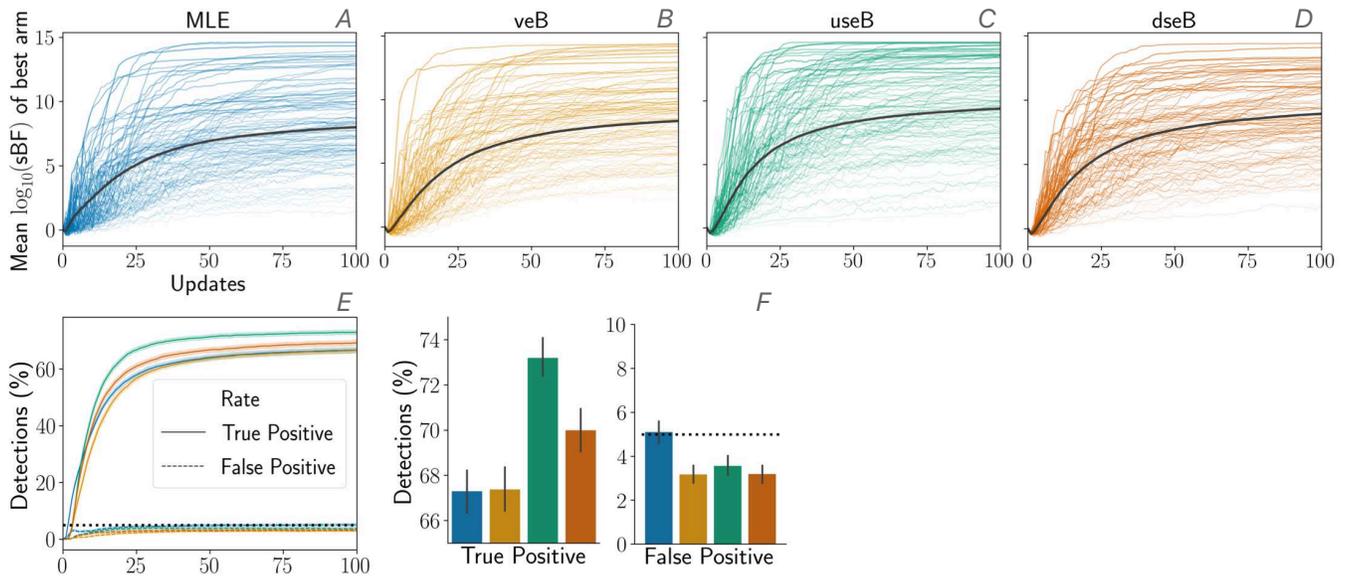}
  \caption{\textbf{Smoothed eB increases statistical power}. In panels A-D, coloured lines represent a sample of 100 individual trials, and plot the mean $\log_{10}$ sequential Bayes factors of the best arm in each trial. Thickness of each line represents mean effect size of the best arm in the corresponding trial. Black line represents mean sBF over all 500 trials. Panel E plots true and false positive rates. Shaded areas represent 95\% confidence intervals, computed over 1000 bootstrap iterations. Panel F plots true and positive rates at the final update in each trial. Horizontal dashed line indicates a 5\% threshold.}
  \label{Stats}
\end{figure*}

Yet another benefit of smoothing becomes evident when evaluating hypotheses about statistically significant differences between treatment arms. As Figs. \ref{Stats}A-D show, the average sequential Bayes factor of the best arm grows more consistently with useB and dseB. In turn, this leads to more accurate and better powered inferential statistics. To demonstrate this, we evaluated Type I and Type II errors by calculating true and false positive rates. Following the recommendation by Deng et al. \cite{Deng2016ContinuousTesting}, we specified an sBF threshold so as to guarantee a false discovery rate of no more than 5\%  with early stopping.

As shown in Fig. \ref{Stats}E, our useB and dseB alternatives achieved higher true positive rates and lower false positive rates. This is brought out clearly in Fig. \ref{Stats}F, which shows that, at the final update, true positive rates were 4-8\% higher and false positive rates were 30-37\% lower with the smoothed eB alternatives than with MLE. The underlying reason for this pattern becomes evident when considering how the smoothed alternatives modulate treatment allocation: because they allocate relatively more treatments to arms other than just the best one, they are better able to model the rewards of these other arms, and thereby significantly improve statistical power.

\subsection{Non-stationary Rewards}
\begin{figure*}[!htbp]
  \centering
  \includegraphics[width=\textwidth]{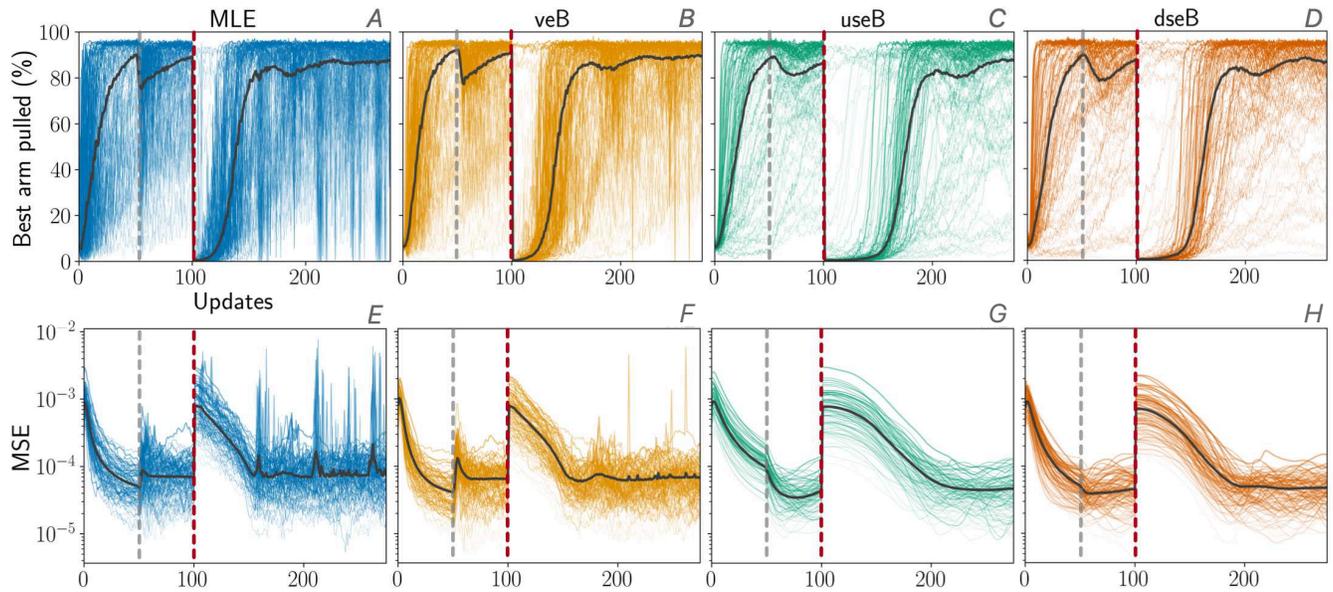}
  \caption{\textbf{Smoothed eB adapts to non-stationary rewards}. Panels A-D plot allocations to the best arm when arm rewards are randomly interchanged at update 100 (dashed red line). Dashed grey line indicates update after which length of smoothing window was bounded to 50 previous updates. Black line represents median allocation percentage to the best arm over all 500 trials. Coloured lines represent 100 individual trials. Thickness of each coloured line represents the effect size between the best arm and the next best one in the corresponding trial. Panels E-H plot MSE. Black line represents MSE averaged over all 500 trials. Thickness of each coloured line represents the variance in true rewards in the corresponding trial.}
  \label{nstat}
\end{figure*}

As described above, delayed feedback causes observed rewards to evolve even if the true rewards themselves do not change. Non-stationarity in the true rewards further compounds the reward estimation problem. We examined a specific case of such non-stationarity where the true rewards are randomly interchanged mid-simulation, at update 100. The length of the smoothing window $U$ was bounded to 50 previous updates, to enable adaptation to this change \cite{Garivier2008OnProblems}. Fig. \ref{nstat} depicts the adaptation of allocations to the best arm and MSE in this scenario. The change in arm rewards and the bounding of $U$ triggered significant ongoing instability in allocations to the best arm in the MLE and veB alternatives (Figs. \ref{nstat}A and \ref{nstat}B), underpinned by greater inter-update variability in reward estimation error (Figs. \ref{nstat}E and \ref{nstat}F). In comparison, useB and dseB evidenced slightly slower but much more stable adaptation in allocations (Figs. \ref{nstat}C and \ref{nstat}D), and much lower inter-update variability in estimation error (Figs. \ref{nstat}G and \ref{nstat}H). As highlighted previously, this speed vs. stability trade-off is controlled by the length and weighting of $U$.

\subsection{Real Rewards and Delays}

\begin{figure}[!htbp]
  \centering
  \includegraphics[width=\columnwidth]{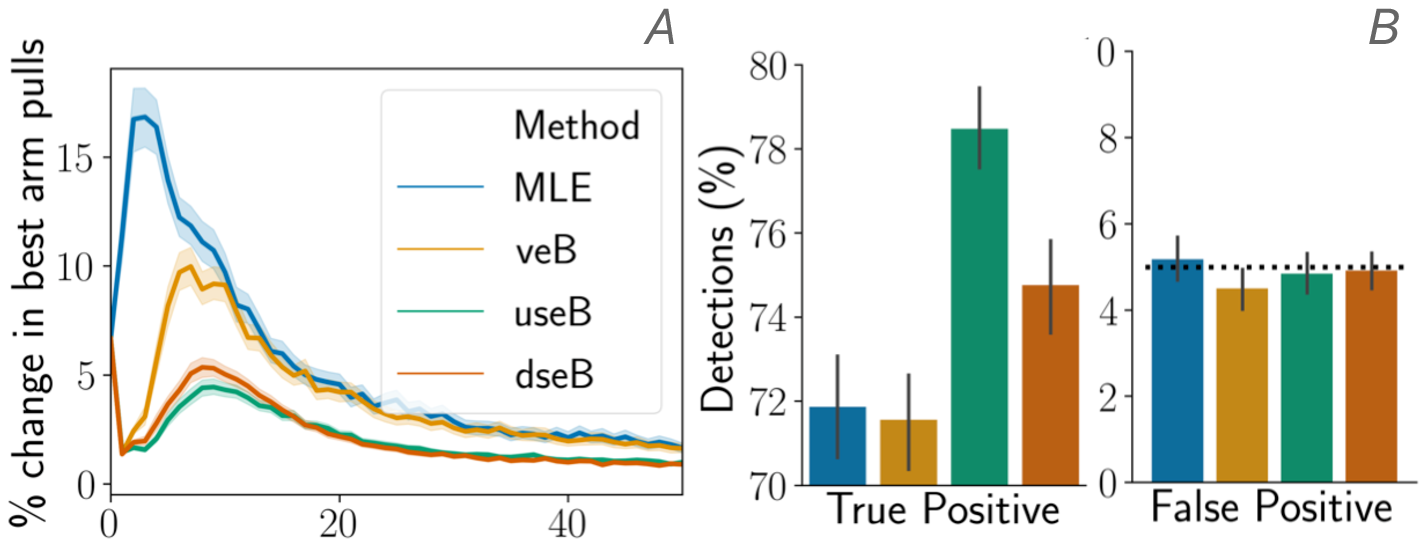}
  \caption{\textbf{Smoothed eB improves performance with real uplifts and delays}. Panel A plots the change in allocations to the best arm. Shaded areas represent 95\% confidence intervals, computed over 1000 bootstrap iterations. Panel B plots true and positive rates at the final update in each trial. Horizontal dashed line indicates a 5\% threshold.}
  \label{Real}
\end{figure}

In this final section, we extend the generality and practical relevance of our results by demonstrating the performance of the smoothed eB alternatives in realistic simulations based on rewards and delays observed at our company (see Figs. \ref{Sims}B and \ref{Sims}C). As with the synthetic simulations, smoothed eB alternatives produced 1.8-2.8x lower peak variability in best arm pulls compared to MLE (compare Figs. \ref{Real}A and \ref{Pulls}E). Further, statistical accuracy was higher, with true positive rates 4-9\% higher than MLE at the final update (compare Figs. \ref{Real}B and \ref{Stats}F).

\section{Conclusions}
This paper has addressed practical challenges in modern adaptive optimisation. Stochastic delays and non-stationarity in feedback are a common phenomenon in online data streams, which necessarily complicate sequential reward estimation. We have proposed computationally lightweight augmentations that in effect, add a further layer of regularisation above empirical Bayesian estimation. The numerical simulations we conducted demonstrate the benefits realised when learning from delayed feedback, in the form of stable adaptation and estimation. These benefits become evident as adaptation progresses alongside changes in true rewards.

We speculate that the benefits our proposal will become more prominent with further increases in the amount and skew in feedback delays, e.g., when treatments include multiple interventions before a response can be generated. Further, there are likely to be other scenarios where it would be valuable. For example, when there are arm-wise asymmetries in the cost of misallocation of treatments. Yet another scenario involves learning with privatised data, where the addition of differential privacy noise could adversely impact accurate reward estimation.

Further, our proposal also benefits the complementary desire for hypothesis testing. This feature is often an important part of \emph{human-in-the-loop} optimisation \cite{Bakshy2018AE:Experimentation}, assisting human controllers in higher-order decision making that is informed by statistically robust inference. Strictly speaking, focusing only on maximising overall reward with adaptive allocation can conflict with reliable hypothesis testing, as few, if any, treatments would be allocated to weaker arms. However, as we show, smoothed empirical Bayesian estimation achieves a balance between these objectives -- not only does it stabilise adaptation with similar regret, it also improves Type I and Type II error rates.

\bibliographystyle{ACM-Reference-Format}
\bibliography{references}

\end{document}